\documentclass[10pt,twocolumn,letterpaper]{article}

\usepackage{cvpr}              

\usepackage{marvosym}
\usepackage{graphicx}
\usepackage{amsmath}
\usepackage{amssymb}
\usepackage{amsfonts}
\usepackage{booktabs}
\usepackage{float}
\usepackage{stfloats}
\usepackage{lipsum}
\usepackage[accsupp]{axessibility}  

\usepackage[pagebackref,breaklinks,colorlinks]{hyperref}

\usepackage[capitalize]{cleveref}
\crefname{section}{Sec.}{Secs.}
\Crefname{section}{Section}{Sections}
\Crefname{table}{Table}{Tables}
\crefname{table}{Tab.}{Tabs.}

\def\cvprPaperID{5110} 
\def\confName{CVPR}
\def\confYear{2022}

\begin{document}

\title{AnyFace: Free-style Text-to-Face Synthesis and Manipulation}

\author{
Jianxin Sun\textsuperscript{1,2}\footnotemark[1], Qiyao Deng\textsuperscript{1,2}\footnotemark[1], Qi Li\textsuperscript{1,2 }\footnotemark[2], Muyi Sun\textsuperscript{1},  Min Ren\textsuperscript{1,2},  Zhenan Sun\textsuperscript{1,2}
\\
\textsuperscript{1} Center for Research on Intelligent Perception and Computing, NLPR, CASIA\\
\textsuperscript{2} School of Artificial Intelligence, University of Chinese Academy of Sciences (UCAS)\\
{\tt\small \{jianxin.sun, dengqiyao, muyi.sun, min.ren\}@cripac.ia.ac.cn, \tt\small \{qli, znsun\}@nlpr.ia.ac.cn}
}



\twocolumn[{
\renewcommand\twocolumn[1][]{#1}
\maketitle
\vspace{-0.8cm}
\begin{center}
\captionsetup{type=figure}
\includegraphics[width=1\linewidth]{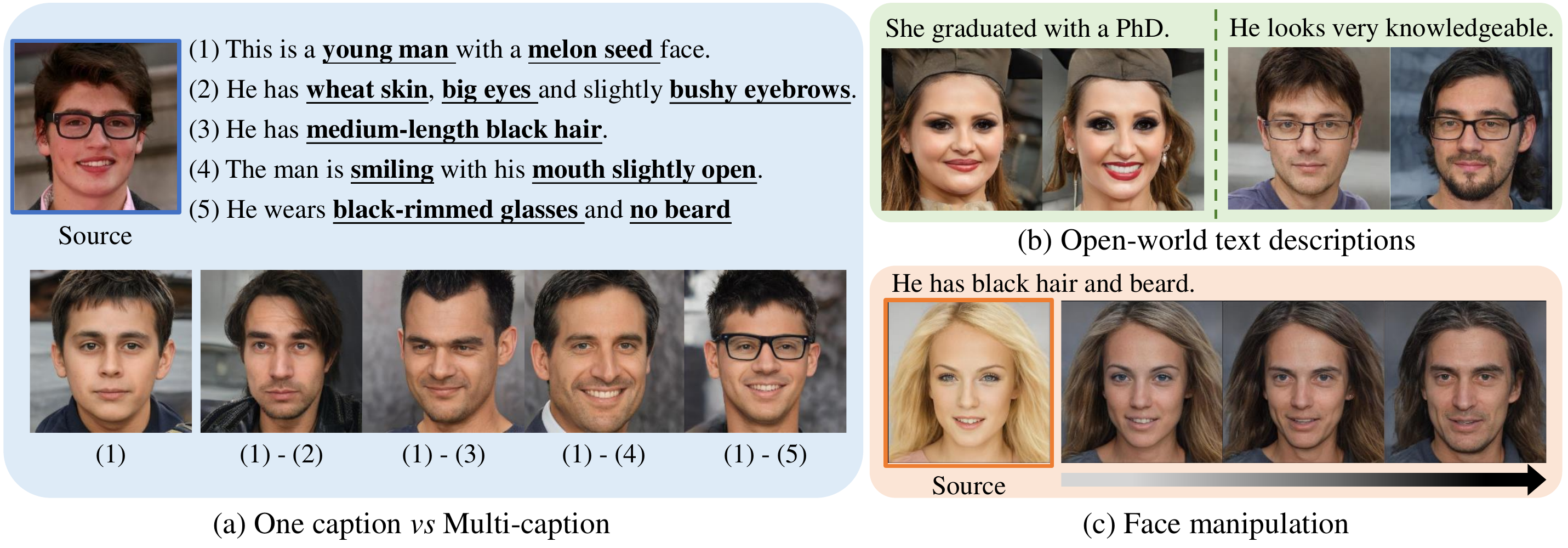}
\vspace*{-3mm}
\captionof{figure}{Our AnyFace framework can be used for real-life applications. (a) Face image synthesis with optical captions. The top left is the source face. (b) Open-world face synthesis with out-of-dataset descriptions. (c) Text-guided face manipulation with continuous control. Given source images, AnyFace can manipulate faces with continuous changes. The arrow indicates the increasing relevance to the text.}
\label{fig1}
\end{center}
}]
\renewcommand{\thefootnote}{\fnsymbol{footnote}}
\footnotetext[1]{Equal contribution}
\footnotetext[2]{Corresponding author}
\begin{abstract}
    Existing text-to-image synthesis methods generally are only applicable to words in the training dataset. However, human faces are so variable to be described with limited words. So this paper proposes the first free-style text-to-face method namely AnyFace enabling much wider open world applications such as metaverse, social media, cosmetics, forensics, etc. AnyFace has a novel two-stream framework for face image synthesis and manipulation given arbitrary descriptions of the human face. Specifically, one stream performs text-to-face generation and the other conducts face image reconstruction. Facial text and image features are extracted using the CLIP (Contrastive Language-Image Pre-training) encoders. And a collaborative Cross Modal Distillation (CMD) module is designed to align the linguistic and visual features across these two streams. Furthermore, a Diverse Triplet Loss (DT loss) is developed to model fine-grained features and improve facial diversity. Extensive experiments on Multi-modal CelebA-HQ and CelebAText-HQ demonstrate significant advantages of AnyFace over state-of-the-art methods. AnyFace can achieve high-quality, high-resolution, and high-diversity face synthesis and manipulation results without any constraints on the number and content of input captions. 
\end{abstract}
\vspace{-3mm}
\section{Introduction}
\label{sec:intro}
\textit{A picture is worth a thousand words. }
Human face, as one of the most important visual signal for social interactions, deserves at least 10,000 words to describe the great diversity in shape, color, hair, expression, and intention, etc. 
Therefore, it is highly desirable to generate variable face images from arbitrary text descriptions for increasing demands of metaverse, social media, cosmetics, advertisement, etc. 
This problem is firstly defined as a free-style Text-to-Face task in this paper because current technology only supports one or two facial captions in the training dataset. 
This paper aims to explore the first plausible solution for free-style Text-to-Face synthesis with successful applications on face image synthesis and manipulation.

Existing face image synthesis~\cite{text2facegan,tedigan,sun2021multi,conditional,yu2019pose,yu2022structure} and manipulation~\cite{ManiGAN,deng2020reference,zhu2021one} methods can synthesize impressive results with the powerful unconditional generative model (e.g., StyleGAN~\cite{stylegan} and NVAE~\cite{vahdat2020nvae}), but vividly generating faces according to specific requirements is still a challenging problem. 
Thus, more and more researchers tend to explore text-to-image (T2I) synthesis \cite{attngan,wang2021cycle,dmgan,rife,han2020cookgan,control,liang2020cpgan,coco}. 
In earlier research, text embedding is directly concatenated with encoded image features and then adopted to generat semantically consistent images by learning the relationship between text and image. 
However, such concatenation operation can only synthesize images from a single caption, limiting the utility of these models in real-life scenarios. 
In practice, one caption can only describe limited information for target images while multiple captions provide fine details and accurate representation. 
As shown in Figure~\ref{fig1}(a), compared with one caption, the face synthesized from 5-caption is more consistent with the source face. 

To handle this issue, several methods~\cite{sun2021multi,rife} attempt to implement multi-caption text-to-image synthesis.
They usually use multi-stage architecture and introduce multi-caption embedding to each stage with special modules. 
Such caption fusion modules extract the features of multiple captions at the cost of external computational resources. 
And more importantly, since an image-text matching network in these methods is pre-trained on the training set, they are still failed in out-of-dataset text descriptions.

Existing text descriptions for image synthesis and manipulation are only limited to a fixed style (\eg fixed number of sentences, formatted grammar, and existing words in datasets), severely limiting the user’s creativity and imagination. 
In real-world applications, any style of text description is more in line with the user's operating needs. Free-style text descriptions include three perspectives: 1) any number of captions (i.e., one or more captions are allowed to describe an image); 
2) any content of captions (allowing users to explore the content or concepts of real-world scenarios); 
3) any format of captions (allowing users to describe an image in their own sentences, rather than following a fixed template).

In this paper, free-style text descriptions are explored, leveraging the power of the recently introduced Contrastive Language-Image Pre-training (CLIP) model and a free-style text-to-face method is proposed namely AnyFace for open-world applications. 
Due to CLIP's ability to learn visual concepts from text-guided natural language, feature extraction of out-of-dataset text descriptions is possible for open-world scenarios (see Figure~\ref{fig1}(b)). 
AnyFace is a two-stream framework that utilizes the interaction between face image synthesis and faces image reconstruction.
The face image synthesis stream generates target images consistent with given texts encoded by the Cross Modal Distillation Module ($\boldsymbol{CMD}$).
The face image reconstruction stream reconstructs the face images paired with the given texts.
These two streams are trained independently, whose mutual information is interacted through a cross-modal transfer loss.
To align the linguistic and visual features, previous attempts~\cite{tedigan,open-world} often leverage a pairwise loss to strictly narrow the distance between text and image features. 
We argue that there is no need to restrict the text embedding to the corresponding image features, as the relationship between text and image is a one-to-many problem. 
The one-to-one constraint may even produce unique results. 
As a remedy, a new Diverse Triplet Loss is proposed to encourage diverse text embedding along with correct visual semantics. 

Overall, the key contributions of this paper are summarized as follows:

\begin{itemize}
    \item To the best of our knowledge, it is the first definition, solution, and application of the free-style text-to-face problem, which is a breakthrough to remove the constraints in face image synthesis. 
    
    \item A novel two-stream framework, which consists of a face synthesis stream and a face reconstruction stream, is proposed for accurate and high fidelity text-to-face synthesis and manipulation. Our method provides a good fusion solution of CLIP visual concepts learning and StyleGAN high-quality image generation. 

    \item Extensive experiments are conducted to demonstrate the advantages of our method in synthesizing and manipulating high fidelity face images. Our work will definitely inspire more creative and wider applications of text-to-face technology. 
\end{itemize}
\vspace{-4mm}
\label{sec:formatting}
\noindent
 \begin{table*}[tbp]
  \centering
  \caption{Comparison of different text-to-image synthesis methods}
  \vspace*{-3mm}
  \label{tab1}
  \renewcommand\tabcolsep{4pt}
  \scalebox{0.95}{
  \begin{tabular}{ccccccccc}
    \toprule
    Methods & AttnGAN~\cite{attngan} & DFGAN~\cite{dfgan}  & RiFeGAN~\cite{rife} & SEA-T2F~\cite{sun2021multi} & CIGAN~\cite{wang2021cycle} & TediGAN-B~\cite{open-world} & \textbf{AnyFace}\\
    \midrule
     Single Model & \checkmark & \checkmark  & \checkmark  & \checkmark  & \checkmark  &- & \checkmark\\
    One Generator & - & \checkmark & - & -  & \checkmark & \checkmark & \checkmark \\
    Multi-caption & - &- &\checkmark &\checkmark  &- &- & \checkmark\\
    High Resolution & - &- & -& -  & \checkmark & \checkmark & \checkmark \\
    Manipulation & - & -& -&  -& \checkmark   & \checkmark & \checkmark\\
    Open-world &  -&- &- &- &- & \checkmark & \checkmark\\
    \bottomrule
  \end{tabular}
  }
  \vspace{-0.2cm}
\end{table*}


\section{Related Works}
\subsection{Text-to-Image Synthesis.}
Currently, multi-stage framework has been widely used for text-to-image synthesis~\cite{attngan,stackgan,stackgan++, dmgan,rife,sun2021multi}. Specifically, sentence features extracted from a pre-trained Bi-LSTM~\cite{bilstm} network are concatenated with noise as input, and multiple generators are applied to synthesize images. However, multiple generators will consume huge computing resources. To address this problem, DFGAN~\cite{dfgan} injects noise by affine transformation~\cite{adain} and uses a single Generator to synthesize images. However, the above method can only generate low-quality images with $256\times 256$ resolution.  
Recently, StyleGAN-based models~\cite{conditional,tedigan,open-world,wang2021cycle}  have been proposed, which advance the traditional methods by a large margin in terms of image quality. These methods usually take StyleGAN~\cite{stylegan} or its followups~\cite{stylegan2,stylegan-ada} as their backbone and train a mapper to project text descriptions into the latent space. Most of them can synthesize high-quality images with resolution at $1024 \times 1024$.
Among them, TediGAN-B~\cite{open-world} attempts to achieve face synthesis with open-world texts, but it requires to re-train a separate model for each sentence and the results are highly random, which hinders its application.
Unlike them, as shown in Table~\ref{tab1}, we can synthesize high-resolution images using a single modal and one generator with multiple and open-world text descriptions.
\subsection{Text-guided Image Manipulation}
Text-guided image manipulation aims at editing images with given text descriptions. Due to the domain gap between image and text, how to map text description to the image domain or learn a cross-modal language-vision representation space is the key to this task. Previous methods~\cite{dong2017semantic,TAGAN} take a GAN-based encoder-decoder architecture as the backbone, and the text embedding is directly concatenated with encoded image features. Such coarse image-text concatenation is ineffective in exploiting text context, thus these methods are limited in manipulated performance. ManiGAN~\cite{ManiGAN} proposes a multi-stage network with a novel text-image combination module for high-quality results. Albeit these methods have explored high-quality results, they are still limited to text descriptions on the dataset. Recently, a contrastive language-image pre-training model (CLIP) ~\cite{clip} is proposed to learn visual concepts from natural language. Based on CLIP's generalization performance, several methods~\cite{open-world, patashnik2021styleclip} achieve face manipulation with out-of-dataset text descriptions. However, these methods require retraining a new model for each given text. Different from all existing methods, we explore a general framework with a single model for arbitrary text descriptions.

\begin{figure*}
  \centering
  \includegraphics[width=1\textwidth]{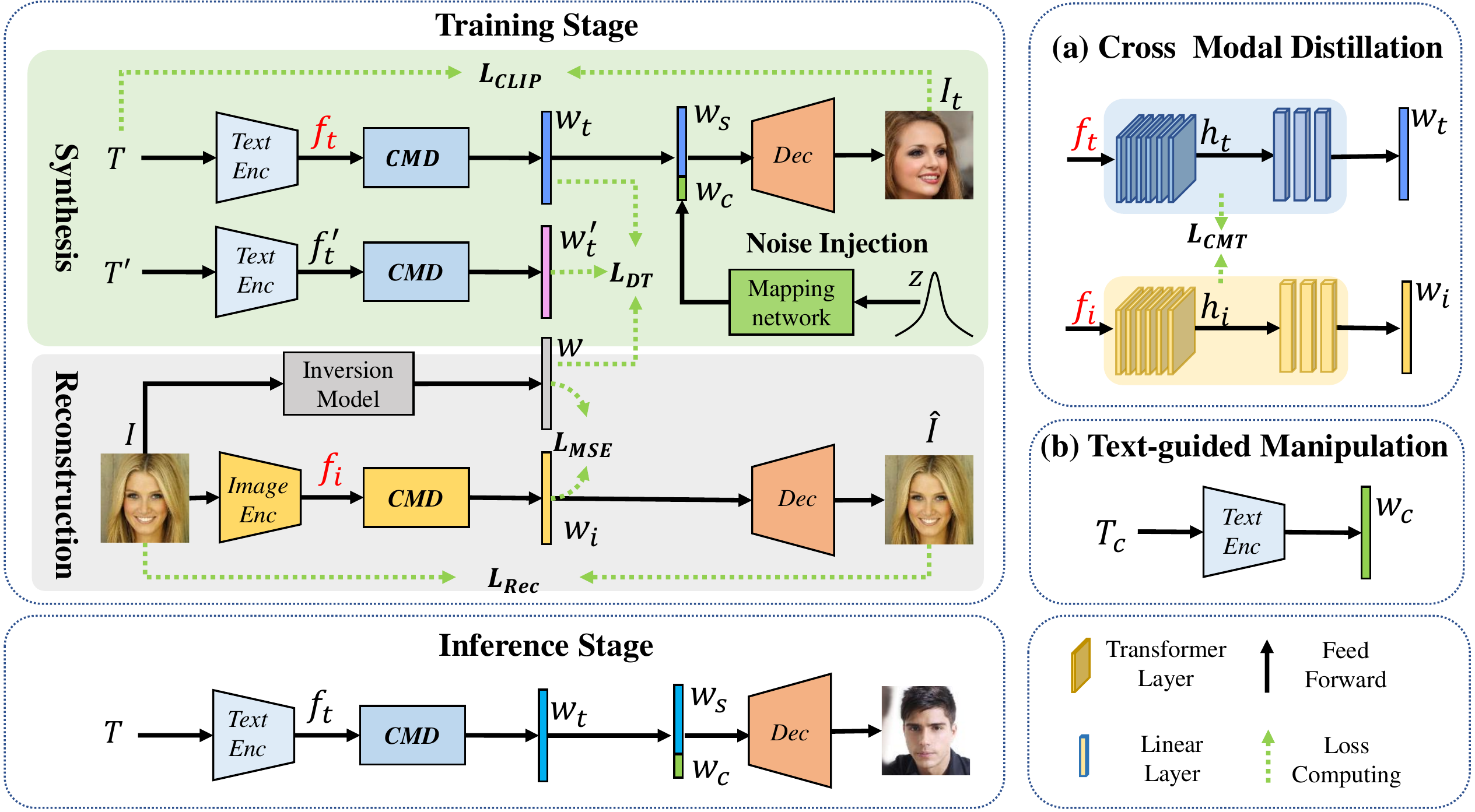}
  \caption{Overview of AnyFace. It consists of a face synthesis network and a face reconstruction network. Text Enc and Image Enc represent CLIP text and image encoder respectively. Dec is the pre-trained StyleGAN decoder. (a) Detailed architecture of Cross Modal Distillation (CMD) module. (b) For text-guided face manipulation, the content code $w_c$ is replaced by the latent code of source image.}
  \label{framework}
  \vspace{-0.4cm}
\end{figure*}

\section{Method}


Figure~\ref{framework} shows an overview of AnyFace, which mainly consists of two streams: the face image reconstruction stream and the face image synthesis stream. 
The training stage is comprised of two streams, while we only keep the face image synthesis stream in the inference stage.
In the following, we will briefly introduce some notations used in this paper. 
Given a face image $\mathbf{I}$ and its corresponding caption $\mathbf{T}$, 
the face image synthesis stream aims to synthesize a face image $\mathbf{I_t}$, whose description is consistent with $\mathbf{T}$. The face image reconstruction stream aims to generate a face image $\mathbf{\hat{I}}$, which can reconstruct the face image $\mathbf{I}$.  To overcome the mode collapse, we also introduce another arbitrary caption $\mathbf{T'}$ and propose a diverse triplet learning strategy for face image synthesis. 


\subsection{Two-stream Text-to-face Synthesis Network}
\label{subsec-twostream}
\paragraph{Face Image Synthesis.}
As shown in Figure~\ref{framework}, this stream is responsible for generating a face image given the input text description $\textbf{T}$. 
We first employ CLIP text encoder to extract a 512 dimensional feature vector $f_t\in{\mathbb{R}^{1\times{512}}}$ from $\textbf{T}$. 
Due to the large gap between linguistic and visual modality, we utilize a transformer based network, namely Cross Modal Distillation ($\boldsymbol{CMD}$), to embed $f_t\in{\mathbb{R}^{1\times{512}}}$ into the latent space of StyleGAN:
\begin{equation}
    w_t=\boldsymbol{CMD}(f_t),
\end{equation}
where  ${w_t} \in {\mathbb{R}^{18 \times 512}}$ denotes the latent code of text features.

Previous works~\cite{patashnik2021styleclip,zhu2021one} have proved that 
the latent space in StyleGAN has been shown to enable many disentangled and meaningful image manipulations.
Instead of generating images by $w_t$ directly, we select the last $m$ dimensions of $w_t$ and denote it as $w_s\in{\mathbb{R}^{m\times{512}}}$. Intuitively, $w_s$ provides the high level attribute information of $\mathbf{I_t}$ learned from the given texts. 
In order to keep the diversity of the generated images, other dimensions of the latent code are harvested from noise by a pre-trained mapping network to provide the random low level topology information. 
We represent it as $w_c \in{\mathbb{R}^{n \times{512}}}$. Unless otherwise specified, we set $m=14$ and $n=4$ in this paper.
Then $w_s$ and $w_c$ are concatenated together and sent to StyleGAN decoder to generate the final result $\mathbf{I_t}$: 
\begin{equation}
    \mathbf{I}_t=\boldsymbol{Dec}\left(w_s\oplus{w_c}\right),
\end{equation}
where $\boldsymbol{Dec}$ denotes StyleGAN decoder, and $\oplus$ represents the concatenation operation.
Note that different with text-to-face synthesis, in the manipulation stage, $w_c$ is generated from the original image using an inversion model to keep the low level topology information of the original image.

\paragraph{Face Image Reconstruction.}

Text-to-face synthesis is an ill-posed problem. 
Given one text description, there may be multiple images that are consistent with it. 
Since the diversity of the generated images is crucial for text-to-face synthesis, we have to carefully design the objective functions to regularize the face image synthesis network so that it can learn more meaningful representations. 
However, directly computing the pixel-wise loss between the original image $\mathbf{I}$ and the synthesis image $\mathbf{I}_t$ is infeasible. 
Thus we design a face reconstruction network, which performs face image reconstruction and provides visual information for the face synthesis network. 

We first exploit CLIP image encoder to extract a 512 dimensional feature vector $f_i\in{\mathbb{R}^{1\times{512}}}$ from   $\mathbf{I}$. 
As shown in Figure~\ref{framework} (a), the reconstructed image can be formulated as:
\begin{equation}
    \mathbf{\hat{I}}=\boldsymbol{Dec}\left(w_i\right),
\end{equation}
where ${w_i} = \boldsymbol{CMD}\left( {{f_i}} \right)\in{\mathbb{R}^{18 \times 512}}$.

Note that both the face synthesis network and
the face reconstruction network have Cross Modal Distillation module, which is important for the interaction between ${w_i}$ and ${w_t}$. 
In the following, we will introduce $\boldsymbol{CMD}$ in detail.

\subsection{Cross Modal Distillation}
Previous works~\cite{mutual,yin2021enhanced,hinton2015distilling} have revealed that model distillation can make two different networks benefit from each other. 
Inspired by them, we also exploit model distillation in this paper. In order to align the hidden features of the face synthesis network and the face reconstruction network and realize the interaction of linguistic and visual information, we propose a simple but effective module named Cross Modal Distillation ($\boldsymbol{CMD}$). 
Recently, the transformer network has achieved impressive performance in natural language processing and computer vision area.

As shown in Figure~\ref{framework} (a), $\boldsymbol{CMD}$ is also a transformer network, which consists of $6$ transformer layers and $3$ linear layers.
Suppose that the output of the transformer layers can be represented as $h_t\in{\mathbb{R}^{1\times{512}}}$
and $h_i\in{\mathbb{R}^{1\times{512}}}$, we should 
align these features before projecting them into the latent space.
Specifically, $\boldsymbol{CMD}$  first normalizes the hidden features by $\rm{softmax}$, and then learns mutual information by the cross modal transfer loss:
\begin{equation}
    \mathcal{L}_{CMT}^{T}=D_{KL}\left(\rm{softmax}(h_t)||(\rm{softmax}(h_i)\right).
\end{equation}
where $D_{KL}$ denotes the Kullbach Leibler(KL) Divergence.
Note that $\mathcal{L}_{CMT}^{T}$ is designed for face synthesis network. We can easily deduce $\mathcal{L}_{CMT}^{I}$ for face reconstruction network. 

Benefiting from the ability of transformer networks to grasp long-distance dependencies, our method can also accept multiple captions as the input without any other fancy modules. To be specific, text features extracted by CLIP text encoder are Stacked as $F_t\in{\mathbb{R}^{n \times 512}}$, and then projected into the latent space by $\boldsymbol{CMD}$ module.

\subsection{Diverse Triplet Learning}

\begin{figure}
  \centering
  \includegraphics[width=1\linewidth]{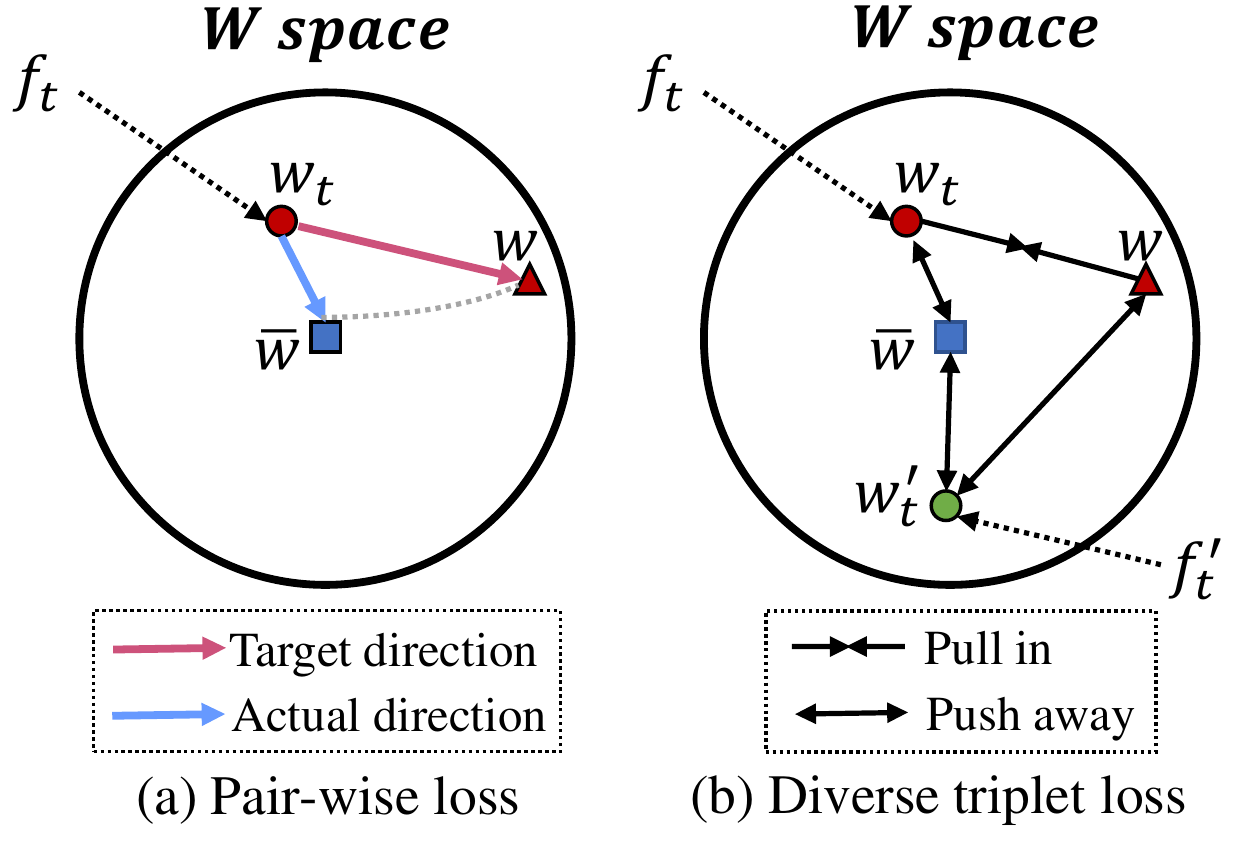}
  \vspace{-2mm}
  \caption{Comparison between (a) pairwise loss and (b) diverse triplet loss. The pairwise loss learns an one-to-one mapping from text embedding $w_t$ to target $w$, but in fact it often converges to average embedding $\overline{w}$ with an unique result. In contrast, diverse triplet loss follows an one-to-many mapping without strict constraints, which encourages positive embedding $w_t$ to be close to $w$ and negative embedding $w'_t$ to stay away from $w$. Meanwhile, both $w_t$ and $w'_t$ stay away from $\overline{w}$. }
  \label{loss}
  \vspace{-0.4cm}
\end{figure}

Recall that in subsection~\ref{subsec-twostream}, $\boldsymbol{CMD}$ embeds the text feature $f_t$ into the latent space of StyleGAN:
$w_t=\boldsymbol{CMD}(f_t)$.
Idealy, $\boldsymbol{CMD}$ reduces the large gaps between the text features extracted by CLIP text encoder and the image features extracted by CLIP image encoder. However, there still exists large gaps between the latent space of CLIP and the latent space of StyleGAN.
A straightforward way to bridge the gap between them is to minimize the pairwise loss:
%

\begin{equation}
    \mathcal{L}_{pair} = \left\|w_t-w\right\|_p 
    \label{pair},
\end{equation}
where $p$ is a matrix norm, such as the
${\ell _0}$-norm, ${\ell _1}$-norm, $w$ is the corresponding latent code of StyleGAN, which is encoded by a pre-trained inversion model~\cite{e4e,psp}.
%
%
However, we find that there is a problem with this  approach.
%
As shown in Figure~\ref{loss} (a), $\boldsymbol{CMD}$ will cheat by converging to the latent code of average face $\overline{w}$, which is possibly close to all of the latent codes in StyleGAN.
%

%

To improve the diversity and synthesize language-driven high fidelity images, we design a \textit{diverse triplet loss}.
First, the latent code of average face $\overline{w}$ is adopted to penalize the model collapse of $\boldsymbol{CMD}$.
Besides, arbitrary caption  $\mathbf{T'}$ is introduced to form a negative pair in face synthesis network. 
The constraints on the positive and the negative pairs are also taken into account:
\begin{equation}
    \mathcal{L}_{DT} = max\left\{\frac{\left \langle w_t,w\right \rangle}{\left \langle w_t,\overline{w}\right \rangle} - \frac{\left \langle w'_t ,w\right \rangle}{\left \langle w'_t,\overline{w} \right \rangle} + m, 0\right\}
    \label{DT},
\end{equation}
where $\left \langle \cdot,\cdot \right \rangle$ refers to the cosine similarity, $m$ represents the margin,  $w$, $w_t^{'}$ correspond to positive image embedding and the negative text embedding respectively.

As shown in Figure~\ref{loss} (b), diverse triplet loss urges positive samples to approach the anchor and negative samples to stay away from the anchor while at the same time encouraging positive and negative samples to diverge and fine-grained features.

\subsection{Objective Functions}

\paragraph{Face Image Synthesis.} The objective function of the face synthesis stream can be formulated as:
\begin{equation}
    \mathcal{L}_{S} = \mathcal{L}_{DT}+\lambda_{CMT}\mathcal{L}_{CMT}^{T}+\lambda_{CLIP}\mathcal{L}_{CLIP},
\end{equation}
where $\lambda_{CLIP}$ is the coefficient of the CLIP loss which is used to ensure the semantic consistency between the generated image and the input text:
\begin{equation}
    \mathcal{L}_{CLIP} = \frac{f_t\cdot{f_{i}^t}} {||f_t||\times||f_{i}^t||},
\end{equation}
$f_t$ and $f_{i}^t$ are features of input text $\mathbf{T}$ and target image $\mathbf{\hat{I}}$.


\paragraph{Face Image Reconstruction.} The objective function of face image reconstruction stream 
can be defined as:
\begin{equation}
    \mathcal{L}_{T} = \mathcal{L}_{MSE}+\lambda_{CMT}\mathcal{L}_{CMT}^{I}+\lambda_{Rec}\mathcal{L}_{Rec},
\end{equation}
where 
$\mathcal{L}_{Rec}$ is employed to to encourage the generated image  $\hat{I}$ to be consistent with the input image $I$, $\mathcal{L}_{MSE}$ is adopted to project the image feature into the latent space of StyleGAN, $\lambda_{Rec}$ is the coefficient for reconstruction loss,  $\lambda_{CMT}$  is the coefficient for  cross modal transfer loss. 
$\mathcal{L}_{Rec}$  and $\mathcal{L}_{MSE}$ are defined as:
\begin{equation}
    \mathcal{L}_{Rec}=||\mathbf{\hat{I}}-\mathbf{I}||_1,
\end{equation}
\vspace{-4mm}
\begin{equation}
    \mathcal{L}_{MSE}=||w_i-w||^{2}_2.
\end{equation}
\vspace{-2mm}


\begin{figure*}[htb]
  \centering
  \includegraphics[width=0.92\textwidth]{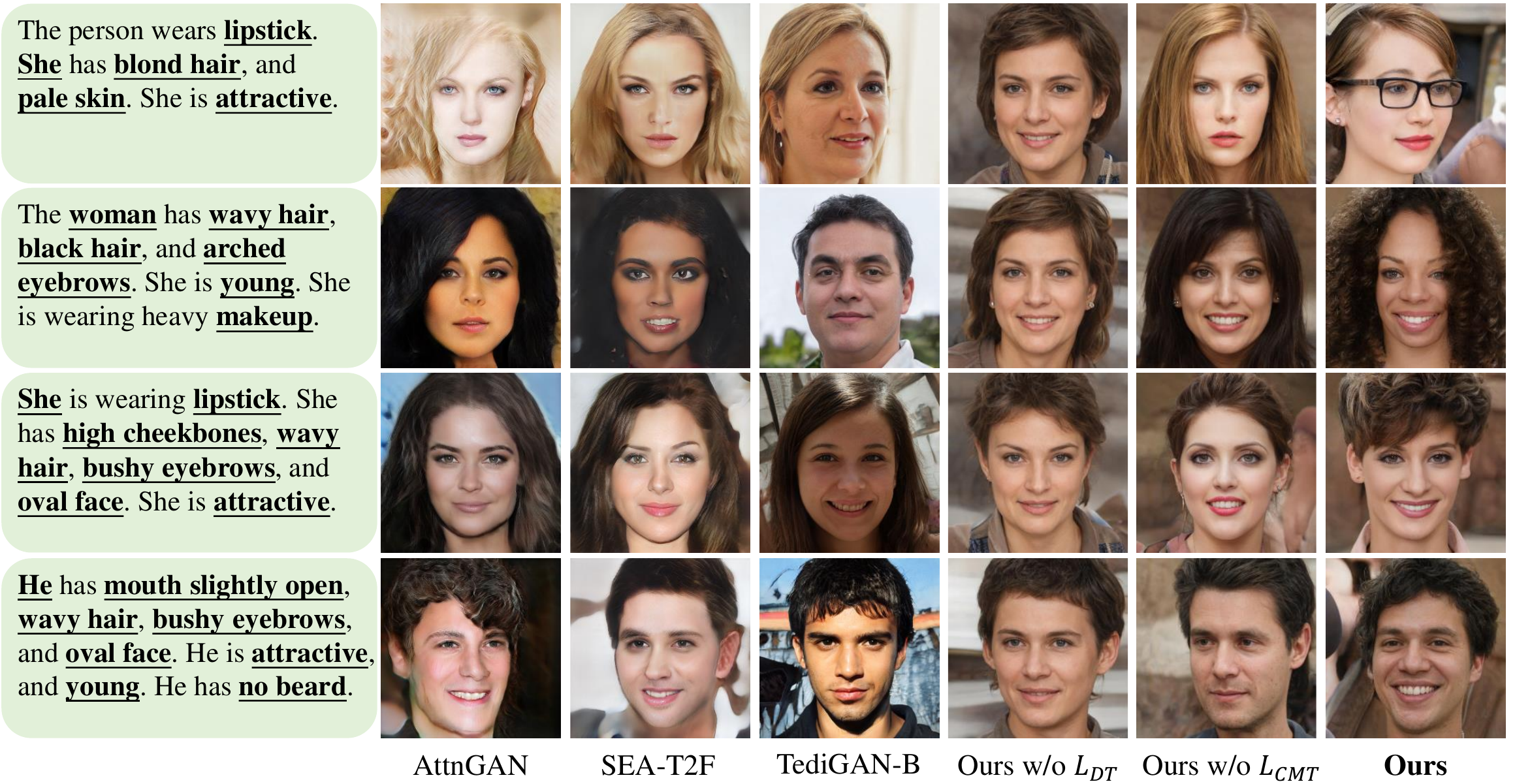}
  \caption{Qualitative comparisons with state-of-the-art methods. The leftmost is the input sentences, columns from left to right represent the results of AttnGAN, SEA-T2F, TediGNA-B, AnyFace without $\mathcal{L}_{DT}$ , AnyFace without $\mathcal{L}_{CMT}$ and our AnyFace respectively.}
  \label{compare}
  \vspace{-0.4cm}
\end{figure*}

\section{Experiments}
\paragraph{Dataset.} We conduct experiments on Multi-Modal CelebA-HQ~\cite{tedigan} and CelebAText-HQ~\cite{sun2021multi} to verify the effectiveness of our method. 
The former has $30,000$ face images with text descriptions synthesized by attribute labels, and the latter contains $15,010$ face images with manually annotated text descriptions. All of the face images come from CelebA-HQ~\cite{CelebAMask-HQ} and each image has ten different text descriptions. We follow the setup of SEA-T2F~\cite{sun2021multi} to split the training and testing set.

\begin{table}[!tbp]
	\caption{Quantitative comparison of different methods on Multi-modal CelebA-HQ dataset. $\downarrow$  means the lower the better while $\uparrow$ means the opposite.}
	\label{eval-multi}
	\vspace{-0.4cm}
	\begin{center}
	\small{
		\begin{tabular}{llll}
			\hline
			Methods & FID $\downarrow$       & LPIPS $\downarrow$   & RFRR $\uparrow$ \\ \hline
			AttnGAN       & 54.22 & 0.548 & 22.15\%    \\
			ControlGAN   &  77.41&  0.572 & 24.5\%    \\
			SEA-T2F & 96.55& 0.545  & 24.3\%   \\ 
			\textbf{AnyFace} &\textbf{50.56} &  \textbf{0.446} &\textbf{29.05\%}  \\
			\hline
		\end{tabular}
		}
	\end{center}
	\vspace{-8mm}
\end{table}
\vspace{-4mm}
\paragraph{Metric.} We evaluate the results of text-to-face synthesis from two aspects: image quality and semantic consistency. Image quality includes reality and diversity, which is evaluated by Fr\'{e}chet Inception Distance (FID)~\cite{fid} and  Learned Perceptual Image Patch Similarity (LPIPS)~\cite{lpips}, respectively. As for semantic consistency, R-precision~\cite{attngan} is used to evaluate traditional T2I methods. An image-text matching network pre-trained on the training set is utilized to retrieval texts for each target image and then calculate the matching rate. One problem of this evaluation metric is that the performance of the pre-trained model is not accurate due to the limited training samples.  Furthermore, there are no pre-trained networks that can match the face and text.
Text-to-face synthesis aims to generate the face image based  on the given text description, 
and we expect the synthesized result to be similar to the original one. Thus, we propose to use the Relative Face Recognition Rate (RFRR) to measure the semantic consistency. Specifically, given the same text description, we use ArcFace~\cite{deng2019arcface} to extract features from images generated by all methods, and calculate the cosine similarity between these features and  the feature of the original image. Then we choose the method with the maximum cosine similarity  as  the successfully matched one.

\begin{table}[!tbp]
	\caption{Quantitative comparison of different methods on CelebAText-HQ dataset. }
	\label{eval-Text}
	\vspace{-0.4cm}
	\begin{center}
	\small{
		\begin{tabular}{llll}
			\hline
			Methods & FID $\downarrow$       & LPIPS $\downarrow$   & RFRR $\uparrow$ \\ \hline
			AttnGAN       & 70.47 & 0.524 & 26.54\%    \\
			ControlGAN   & 91.92 & 0.512 & 25.85\%     \\
			SEA-T2F & 140.25 &  0.502 & 18.31\% \\ 
			\textbf{AnyFace} &\textbf{56.75}& \textbf{0.431}  & \textbf{29.31\%}  \\
			\hline
		\end{tabular}
		}
	\end{center}
	\vspace{-6mm}
\end{table}

\subsection{Quantitative Results}
We compare our AnyFace with previous start-of-the-art models on CelebAText-HQ and Multi-modal CelebA-HQ dataset.
As shown in Table~\ref{eval-multi}, on Multi-Modal CelebA-HQ dataset, AnyFace outperforms the current best method with an improvement of $3.66$ FID, $0.099$ LPIPS, and $4.55\%$ RFRR points.
More impressively, AnyFace reduces the best reported FID on the CelebAText-HQ dataset from $70.49$ to $56.75$, a $19.5\%$ reduction relatively.
The CelebAText-HQ is much more challenging because it is manually manuated, but we can also synthesize  high reality, diverse and accurate face images given the text description.

\begin{figure*}
  \centering
  \includegraphics[width=0.9\textwidth]{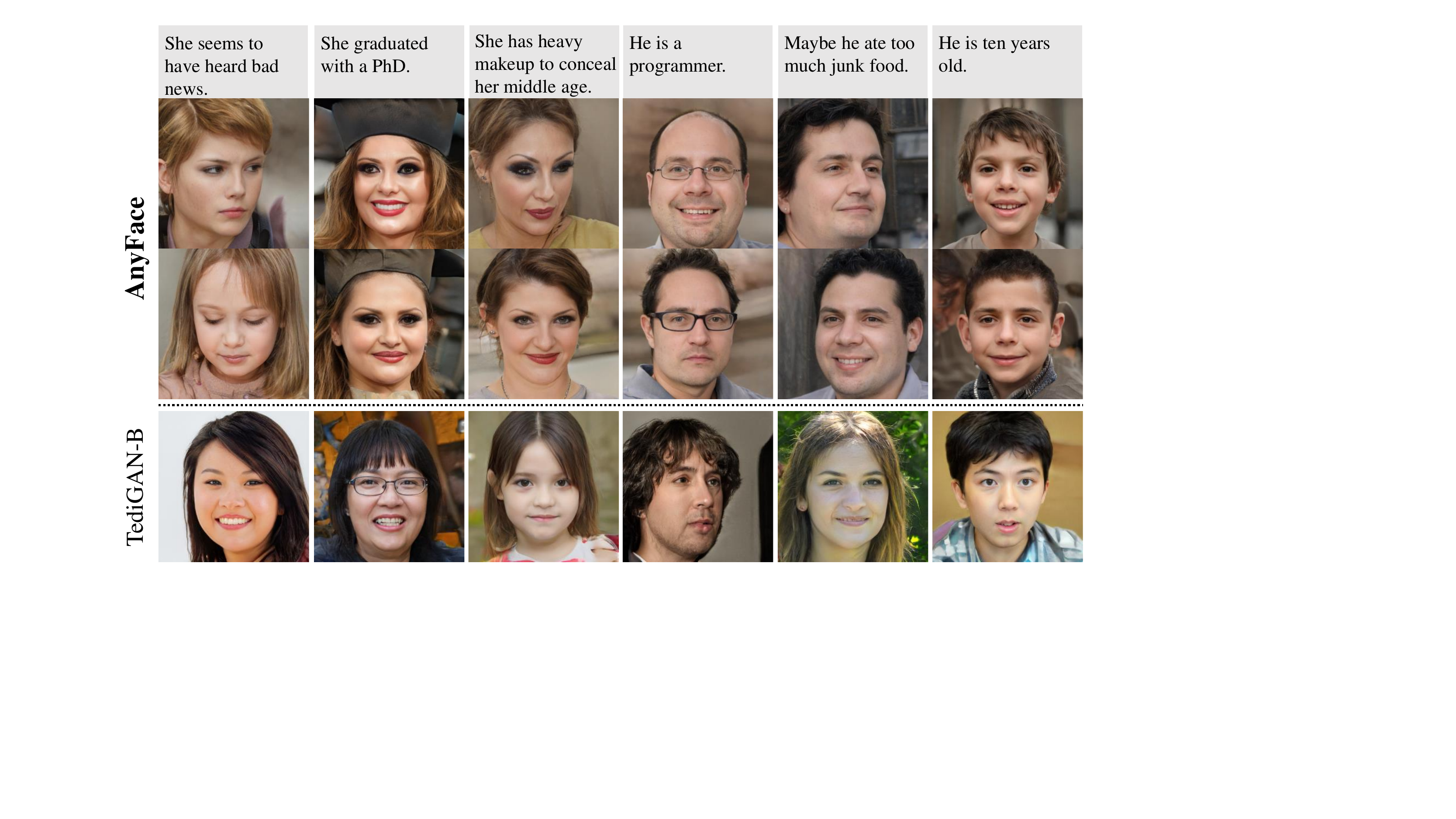}
  \caption{Illustrations of open-world text-to-face synthesis. The first row represents the text descriptions harvested from the Internet, and the rest of column are generated results with given text.}
  \label{open}
  \vspace{-0.4cm}
\end{figure*}

\subsection{Qualitative Results}
In this subsection, we compare our results with other state-of-the-art methods qualitatively.
Moreover, we present the generalization of the proposed method in real-life applications, including open-world, multi-caption and text-guided face manipulation scenarios.
\vspace{-2mm}
\subsubsection{Qualitative Comparisons}
Both image quality and semantic consistency should be considered when evaluating the results of different T2I methods qualitatively. 
As shown in Figure~\ref{compare}, though multi-stage method (AttnGAN~\cite{attngan} and SEA-T2F~\cite{sun2021multi})  can synthesize images corresponding to part of the text descriptions, they can only generate low quality images with the resolution at $256\times256$.
While TediGAN-B~\cite{open-world} successfully synthesizes images with high quality. However, the results seem to be text-irrelevant. For example, in the second row, it generates a male image while it should be a female according to the text description.
Our method introduces  $\boldsymbol{CMD}$ module and Diverse Triplet loss to help the network to learn some fine-grained features such as beard, smile, makeup, mouth type, hair color and shape, etc., which illustrates that the images synthesized by our method are much better in terms of image quality and semantic consistency. 
\vspace{-2mm}

\subsubsection{Real-life Applications}

\textbf{Open-world Scenarios.} 
In this subsection, we compare with TediGAN-B~\cite{open-world} for open-world scenarios. 
TediGAN-B adopts instance level optimization for open-world texts. Thus it requires retraining the network for each text description, which is tedious and time-consuming. 
As for our method, a single model is trained on one dataset and can be easily extend to unseen text descriptions.

As shown in Figure~\ref{open}, TediGAN-B can generate photo-realistic results. However, these results are almost text-irrelevant. 
While our method can not only synthesize vivid images which are text-relevant, but can generate images with more diversity due to the sophisticated design of the face synthesis network.
Besides, benefiting from the powerful representation of CLIP, text descriptions with similar meanings will be encoded into similar latent codes of StyleGAN even though the text descriptions contains abstract information. 
Overall, AnyFace learns visual concepts from text descriptions. For example, "bad news" produces the expression ``sad'', ``PhD'' corresponds to ``mortarboard'', and ``programmer'' is related to ``bald'' and ``glasses''.

\begin{figure}
  \centering
  \includegraphics[width=0.92\linewidth]{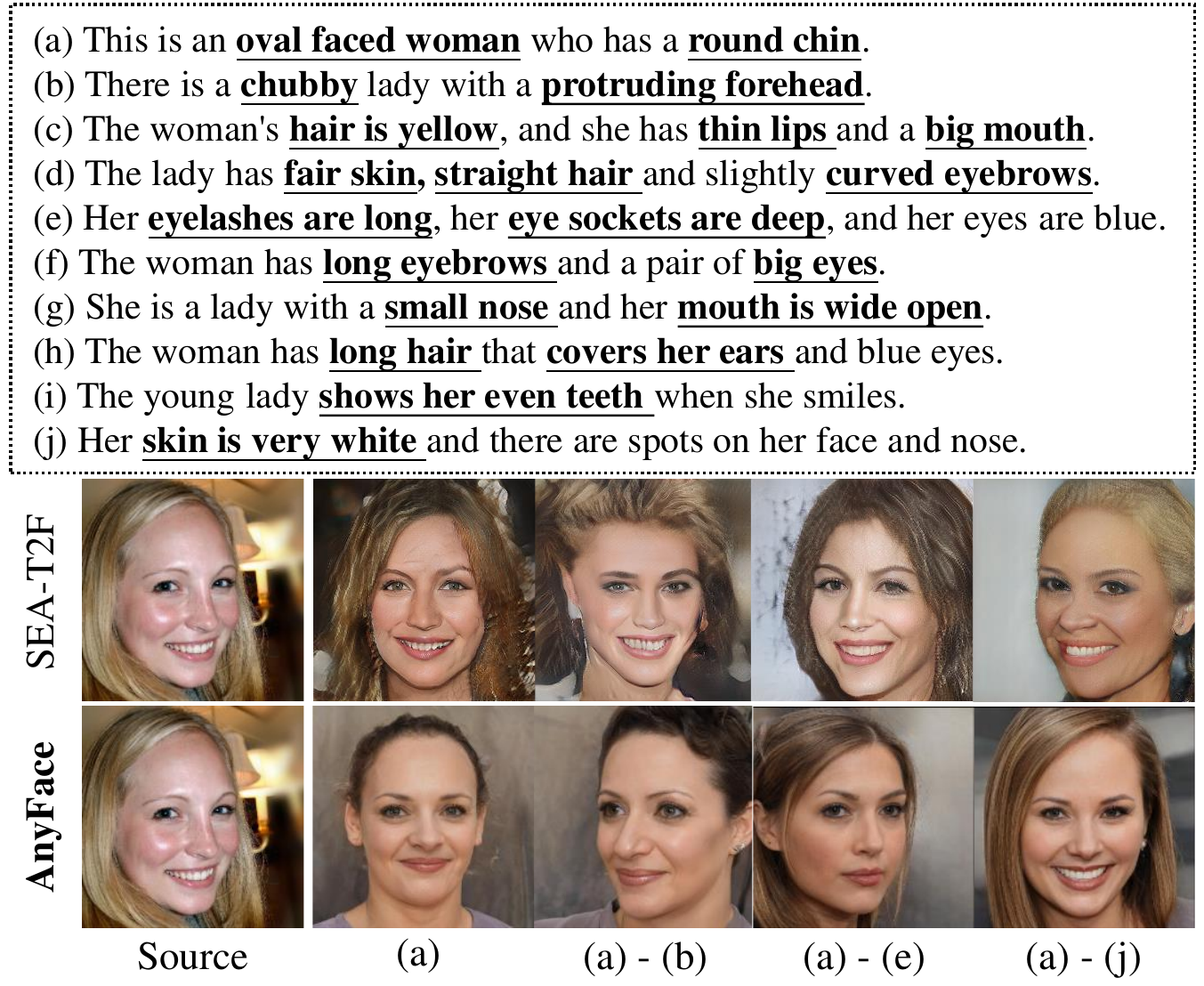}
  \caption{Illustrations of multi-caption synthesis. The text description above corresponds to the source image, synthesized results are conditioned by (a), (a-b), (a-e) and all captions, respectively.}
  \label{mul}
  \vspace{-0.4cm}
\end{figure}
\vspace{0.1cm}
\textbf{Multi-caption Scenarios.}
In this subsection, we compare with SEA-T2F~\cite{sun2021multi} for multi-caption scenarios. 
As shown in Figure~\ref{mul}, AnyFace produces natural and smooth appearance without obvious ghosting artifacts and color distortions. 
As the number of sentences increases, the difficulty of matching all texts correctly also increases. Compared with SEA-T2F, the results of AnyFace are more consistent with the text descriptions (e.g., ``long hair'' and ``curves her ears'') . Note that due to the subjectivity of the labeling data samples in~\cite{sun2021multi}, even the source images may not perfectly match all the descriptions.
\vspace{0.1cm}

\begin{figure}
  \centering
  \includegraphics[width=0.92\linewidth]{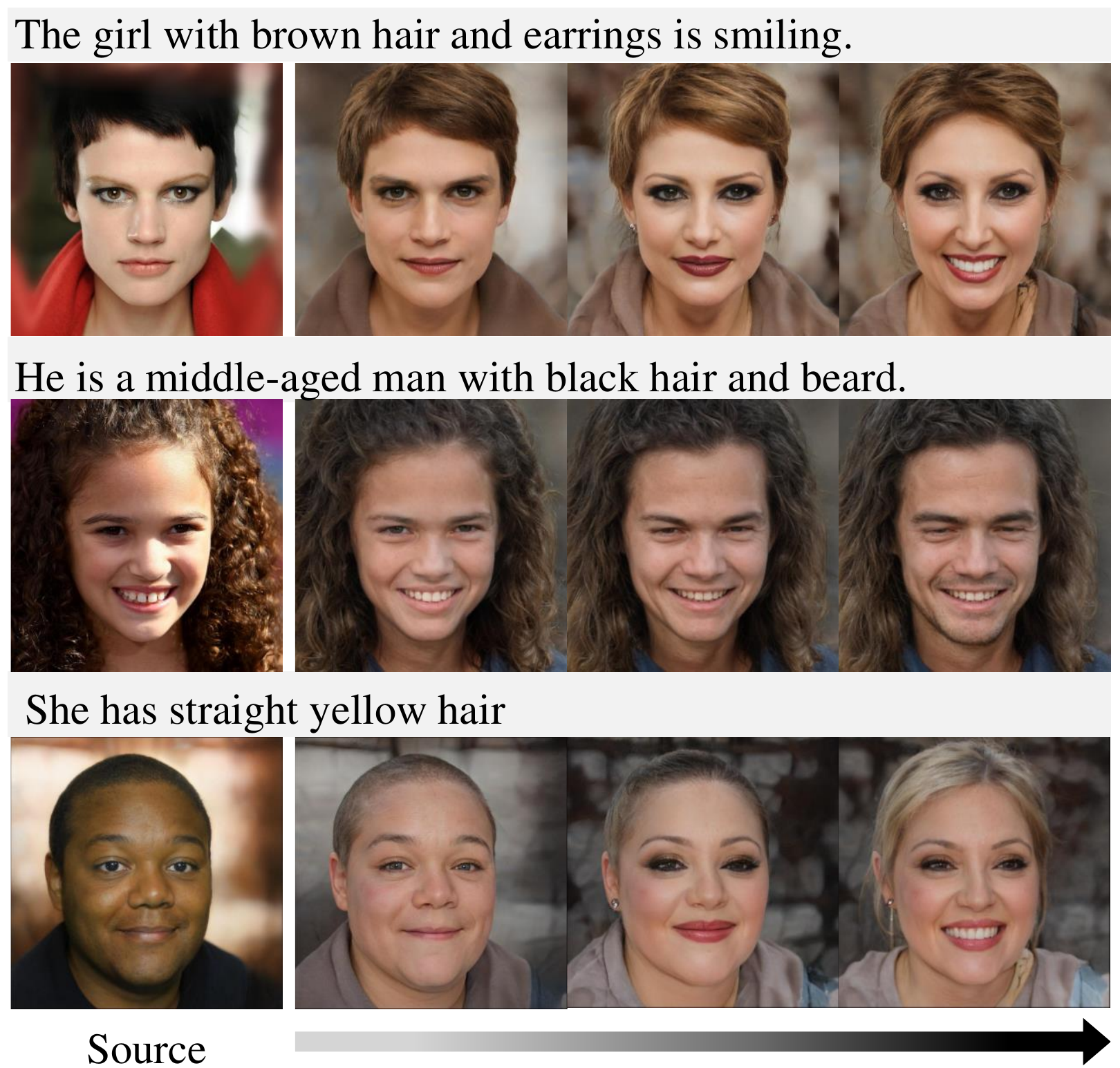}
  \caption{Illustrations of text-guided face manipulation with continuous control. Given source images (1st column), manipulated images show continuous changes according to the text. The arrow indicates the increasing relevance to the text}
  \label{mani}
  \vspace{-0.4cm}
\end{figure}

\textbf{Text-guided Face Manipulation.}
In this subsection, we show that AnyFace can also  be easily adapted to manipulate face images continuously given the text description by changing the size of text embedding (i.e., changing $m$ and $n$ in subsecton~\ref{subsec-twostream}). 
From Figure~\ref{mani}, we observe that AnyFace can manipulate faces with any text descriptions, including global (e.g., smiling), local (e.g., brown hair) and abstract (e.g., middle-aged) attributes. Meanwhile, as the text information increases, the generated face will be more consistent with the text information.
\vspace{-1mm}

\subsubsection{Ablation Study.} 
In this part, we conduct ablation studies to evaluate the contributions of diverse triplet loss and cross-modal transfer loss to our framework. 
As shown in  Figure~\ref{compare} (``w/o $\mathcal{L}_{DT}$ ''), we replace the diverse triplet loss with the pairwise loss. Synthesized faces tend to have similar appearance, such as smile, short hair and natural skin. In contrast, AnyFace with $\mathcal{L}_{DT}$ produces diverse and personalized characteristics, e.g., different hairstyles, accessories, skin tones, etc. 

\begin{figure}
  \centering
  \includegraphics[width=0.92\linewidth]{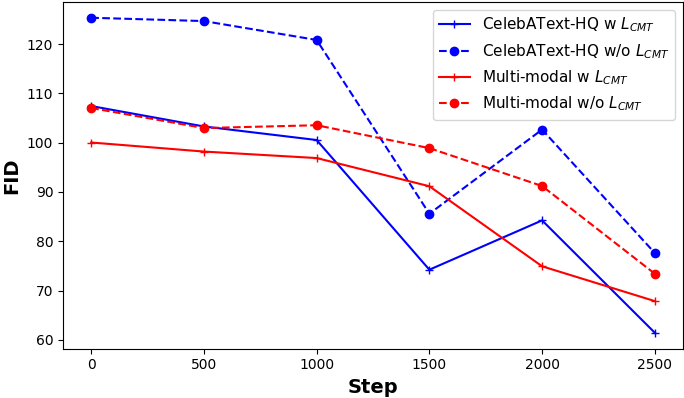}
  \caption{FID of AnyFace with or without $\mathcal{L}_{CMT}$ in different steps.}
  \label{CMT}
  \vspace{-0.4cm}
\end{figure}

We further present the impact of the cross-modal transfer loss on our framework by comparing the qualitative results with or without $\mathcal{L}_{CMT}$ in Figure~\ref{compare}. It is difficult for ``AnyFace w/o $\mathcal{L}_{CMT}$ '' to learn structured features. For example, faces of the second and third rows fail to synthesize `` wavy hair'', and the synthesized mouth in the last row is not consistent to part of the text description: ``mouth slightly open''. The effect of $\mathcal{L}_{CMT}$ is further explored from a quantitative perspective. In Figure~\ref{CMT}, we demonstrate that FID changes with or without $\mathcal{L}_{CMT}$ as training steps increases. We observe that the solid line (``w $\mathcal{L}_{CMT}$'') improves FID by a large margin and speeds up convergence on all of the  datasets. Both quantitative and qualitative assessments demonstrate the effectiveness of proposed losses.
\vspace{-2mm}

\vspace{-2mm}
\section{Limitation}

Our method has three limitations. 1) Our method cannot infer the identity information from the  text descriptions, such as `Donald Trump'; 2) The results generated by the same text description have a similar style; 3) Sometimes, the attributes that are irrelevant to the text will also be changed during manipulation process.

\section{Conclusion}
In this paper, we propose a two-stream framework for text-to-face synthesis, which consists of a face synthesis stream and a face reconstruction stream.
A Cross Modal Distillation module is further introduced to align the information between the two streams and a diverse triplet loss helps the network to produce images with diverse and fine-grained face components. 
Furthermore, our method can be applied to real-world scenarios such as open-world or multiple caption text-guided face synthesis and manipulation.

\vspace{-1mm}
\noindent 
\paragraph{Acknowledgements:}This work was supported in part by the National Key Research and Development Program of China under Grant No. 2020AAA0140002, in part by the Natural Science Foundation of China (Grant Nos. U1836217, 62076240).
{\small
\bibliographystyle{ieee_fullname}
\bibliography{egbib}
}

\end{document}